\documentclass[journal,12pt,onecolumn,draftclsnofoot,]{IEEEtran}

\usepackage{cite}
\usepackage{amsmath,amssymb,amsfonts}
\usepackage{algorithm}
\usepackage{algpseudocode}
\usepackage{graphicx}
\usepackage{textcomp}
\usepackage{xcolor}
\usepackage{listings}
\usepackage{footnote}
\usepackage[hidelinks]{hyperref}
\usepackage[justification=centering]{caption}
\usepackage{setspace}
\usepackage[T1]{fontenc}
\usepackage{times}
\usepackage{wrapfig}
\usepackage{tabto}
\usepackage{derivative}
\usepackage{amsmath}
\usepackage{mathtools}
\usepackage{caption}
\usepackage{subcaption}
\usepackage{multicol}
\usepackage{fancyhdr}
\usepackage{blindtext}
\usepackage{lipsum}
\usepackage{centernot}
\usepackage{bbm}
\usepackage{calrsfs}

\DeclareMathAlphabet{\pazocal}{OMS}{zplm}{m}{n}

\def\BibTeX{{\rm B\kern-.05em{\sc i\kern-.025em b}\kern-.08em
    T\kern-.1667em\lower.7ex\hbox{E}\kern-.125emX}}

\title{\vspace*{\fill}Optimizing LaneSegNet for Real-Time Lane Topology Prediction in Autonomous Vehicles\\ {\large Purdue University, West Lafayette, Indiana\\Image Processing and Analysis Lab\\May 2024}\vspace*{\fill}}
\author{\raggedright Primary Author: William Stevens\\
Co-authors: Vishal Urs, Karthik Selvaraj, Gabriel Torres, Gaurish Lakhanpal\\
Mentors: Professor Edward J. Delp and Professor Carla B. Zoltowski}
\date{}

\usepackage{fancyhdr}
\begin{document}
\maketitle
\thispagestyle{empty}

\newpage
\pagestyle{fancy}
\tableofcontents

\newpage 
\section{\textbf{Introduction}}
As autonomous vehicles become increasingly prevalent, it is crucial to ensure the computer software powering these vehicles is able to make intelligent and safe driving decisions. These decisions are informed by real-time assessments of road conditions through computer vision algorithms. Through Purdue’s Vertically Integrated Projects (VIP) program, and under the mentorship of Dr. Edward J. Delp and Dr. Carla B. Zoltowski, whose knowledge, guidance, and support have been instrumental in our work, our team set out to contribute to this cause by investigating one of the recent techniques introduced to the field. Our machine learning development group represents the Lane Detection team, within the larger Image Processing and Analysis (IPA) cohort, which has been striving to research, implement, and test computer vision techniques for autonomous driving purposes. Each semester, current members have set out to build on learning achieved by previous teams, with the most recent culmination of this coming through our team’s work in the Spring term of 2024.

The goal of this semester was to expand upon the computer vision algorithms investigated in the previous semester, specifically pivoting towards obtaining an output that was more useful to autonomous driving software. Though the Fall 2023 team was successful in its development of a promisingly accurate Detection Transformer (DETR) \cite{detr} architecture, we realized that DETR’s output of classified bounding boxes was not viable to autonomous driving software. Thus, the main objective of the team was to research, implement, and test a new computer vision framework that would take dashboard images and transform them into useful semantic information.

\newpage 
\section{\textbf{Methodology}}

\begin{figure}[H]
    \centering
    \includegraphics[width=1\columnwidth]{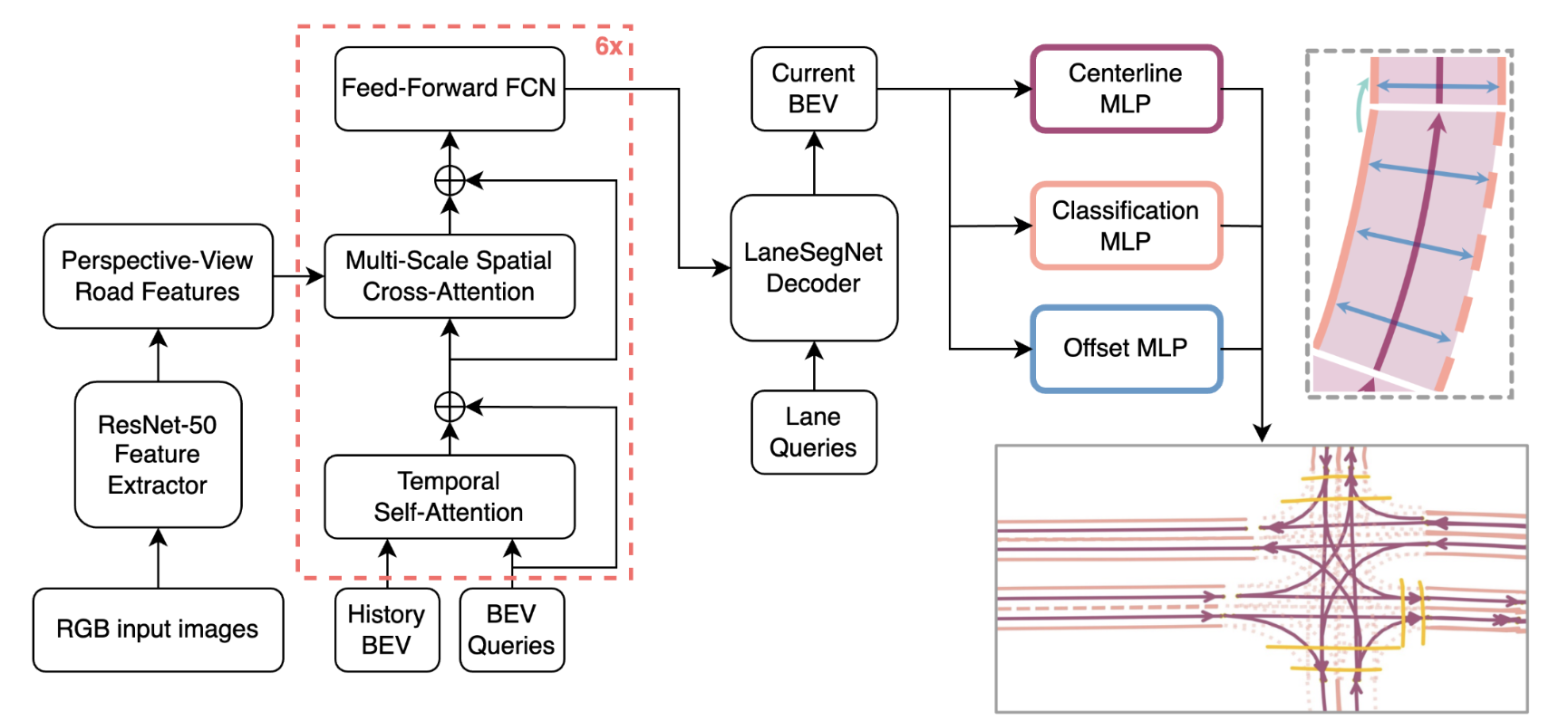}
    \caption{LaneSegNet Architecture.}
    \label{fig:lanesegnet_arch}
\end{figure}

Building off of the DETR results from the previous semester, the team investigated the LaneSegNet \cite{lanesegnet}  architecture. LaneSegNet was first introduced in December 2023 by Li et al. and leverages the OpenLane-V2 \cite{openlanev2} dataset to build a lane topology prediction model. LaneSegNet’s model provides more accurate predictions than other autonomous driving frameworks due to its ability to integrate topological information with lane-line information, resulting in a more defined and contextual understanding of the geographic environment. Figure \ref{fig:lanesegnet_arch} illustrates the LaneSegNet architecture, which incorporates components from other networks like ResNet-50 \cite{ResNet}, BEVformer \cite{bevformer}, and various attention mechanisms. The different attention mechanisms are all derived from the classic attention mechanism proposed in  "Attention is All You need" \cite{attention}, the novel paper that first introduced transformers as a machine learning technique. The architecture of LaneSegNet can be understood through four main stages: the feature extractor, the lane encoder, the lane decoder, and the prediction head. Joined together, these components will produce the model’s prediction, based on its learned understanding of the input images and history predictions.

\newpage 

\begin{figure}[H]
\centering
\begin{subfigure}{.5\textwidth}
  \includegraphics[width=1.5\linewidth]{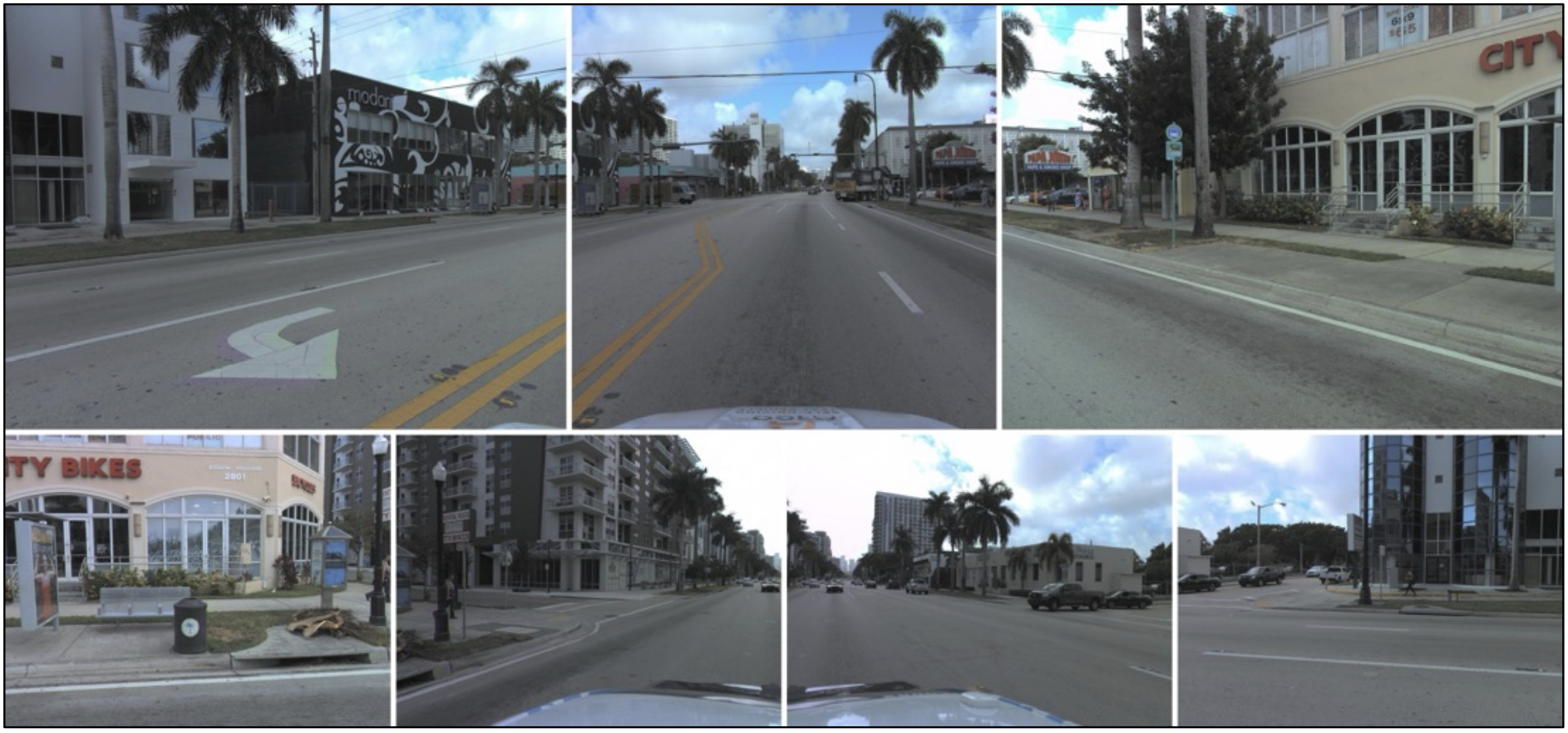}
  \caption{\raggedleft Surrounding View Images}
  \label{fig:openlaneimages}
\end{subfigure}%
\begin{subfigure}{.5\textwidth}
  \raggedleft
  \includegraphics[width=.4\linewidth]{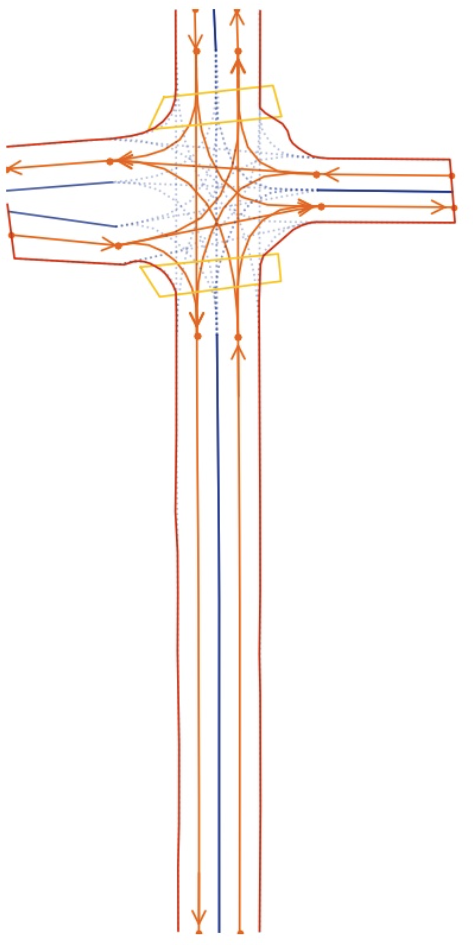}
  \caption{\raggedleft Ideal Output}
  \label{fig:openlaneannotation}
\end{subfigure}
\caption{Example Images and Output from OpenLane-V2. \cite{openlanev2}}
\label{fig:openlane}
\end{figure}

LaneSegNet, as proposed by its authors, is designed to accept seven camera angles covering a 360$^o$ span around the car (Figure \ref{fig:openlaneimages}). These images are propagated through the model architecture, yielding a drawn diagram representing the model’s prediction of a satellite view of what the road scenario looks like (Figure \ref{fig:openlaneannotation}). These predictions are further strengthened by LaneSegNet’s unique design, which regularly updates the lane topology output by incorporating its prediction history into the current prediction.

In an attempt to contribute to the work achieved by the LaneSegNet authors, the team conducted a design study of the model's underlying architecture in an attempt to deepen our understanding of the intuition behind its design and to search for ways to optimize or improve its performance. After researching and discussing the concepts of LaneSegNet to ensure we understood the reasoning and intuition behind its design, we set out to implement the model in code, closely following the official source code uploaded by the authors on GitHub \cite{lanesegnetgithub}. In an attempt to recreate the impressive accuracy obtained by the authors' implementation, we trained the model from scratch on the OpenLane-V2 dataset using a singular NVIDIA Tesla A100 (80GB) GPU. We were extremely fortunate to be granted access to use this powerful GPU through our connection to Dr. Lei Wang's research laboratory at Purdue. The official LaneSegNet implementation was trained on eight NVIDIA Tesla A100 (32GB) GPUs in parallel, allowing for roughly six times faster training than our GPU, according to the Lambda Labs benchmark comparison \cite{Balaban_2023}. As such, it is evident that successful training of the LaneSegNet architectures requires an extremely high level of computational power, due to the model's size and complexity. One of the more prominent motivators of this design study was to find hyperparameter adjustments that would allow users to train the model on much more affordable and accessible processors. To have access to even one high-powered GPU, let alone eight, is a very exclusive privilege. Therefore, the methods carried out in this design study are meant to provide users with viable procedures for training LaneSegNet at a cheaper computation, cost without sacrificing too much accuracy.

\section{\textbf{Optimization Methods}}

\subsection{Feature Extractor Optimization}
Our first attempt at reducing the computational complexity of LaneSegNet was to reduce the number of convolutional layers in its feature extractor. As mentioned previously, the official implementation uses ResNet-50 \cite{ResNet}, containing 50 convolutional layers and residual connections. The purpose of this component is to examine the images taken from the seven cameras around the car, and perform sequential convolutions in order to extract the important features present. Over time, the network will learn to extract features that represent the inherent patterns and structure that are relevant to the semantic information we want to extract. In convolutional neural networks, high level features extracted further in the network are often low-resolution edge patterns, and lower level features represent certain combinations of such edges. In autonomous driving images, such features help provide a more complete semantic understanding of image attributes such as crosswalks or curbs. These features, representative of the road scenario from the ground level, are what enhance the lane encoder by allowing it to transform the ground level perspective view (PV) to the aerial birds-eye-view (BEV).

\begin{figure}[H]
    \centering
    \includegraphics[width=0.9\columnwidth]{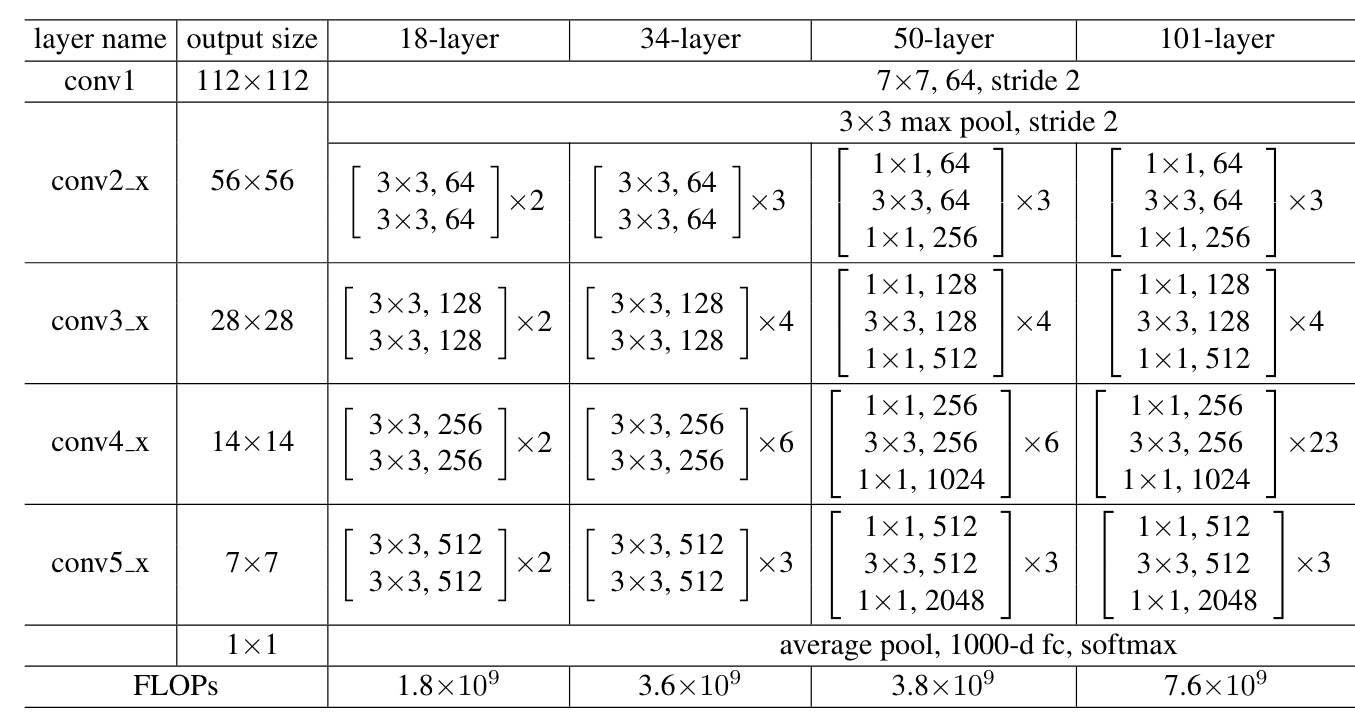}
    \caption{ResNet Models: FLOPs Comparison. \cite{ResNet}}
    \label{fig:resnet_flops}
\end{figure}

According to the official experimentation conducted by Kaiming, et al. in the original paper introducing ResNet entitled "Deep Residual Learning for Image Recognition" \cite{ResNet}, training various residual networks on ImageNet found that the 50-layer network requires 3.8 billion floating-point operations (FLOPs), whereas a compressed version of the network containing only 18 layers uses less than half the computation, at 1.8 billion FLOPs (Figure \ref{fig:resnet_flops}). This reduction in complexity attracted the project team towards the possibility of replacing LaneSegNet's ResNet-50 feature with ResNet-18, to save training time. 

\begin{figure}[H]
    \centering
    \includegraphics[width=0.4\columnwidth]{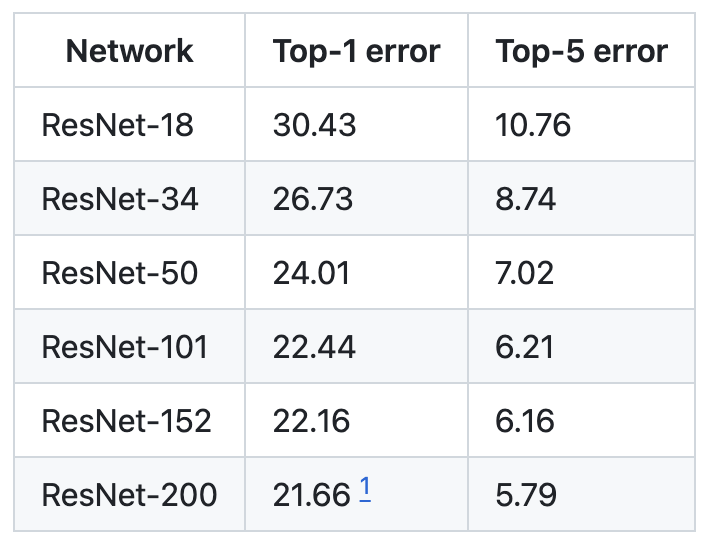}
    \caption{ResNet Models: Error Comparison. \cite{Facebookarchive_2016}}
    \label{fig:resnet_accuracy}
\end{figure}

The only consideration left was how using a shallower feature extractor would impact the model's accuracy. Further investigation led the team toward the Facebook benchmark for ResNet models \cite{Facebookarchive_2016}. This benchmark is shown in Figure \ref{fig:resnet_accuracy}, and illustrates how ResNet-18 has just a $6\%$ higher error for Top-1 classifications on ImageNet, and a $4\%$ higher error on Top-5 classifications. This slight drop in accuracy from ResNet-50 to ResNet-18 therefore led us to believe that it was worth investigating how such a change would impact LaneSegNet in terms of computation time and accuracy.

\subsection{Encoder and Decoder Optimization}

Another optimization the team investigated in this study was through hyperparameter tuning, specific to the encoder and decoder in the LaneSegNet design. The classic transformer design, first introduced in "Attention is All You need" \cite{attention}, uses a special equation called an attention mechanism to compute the inherent connections between tokens in an input sequence, in the form of an attention matrix. Variations of the original attention mechanism are used in the encoder and decoder components of all transformer models, including LaneSegNet.

\begin{figure}[H]
    \centering
    \includegraphics[width=0.8\columnwidth]{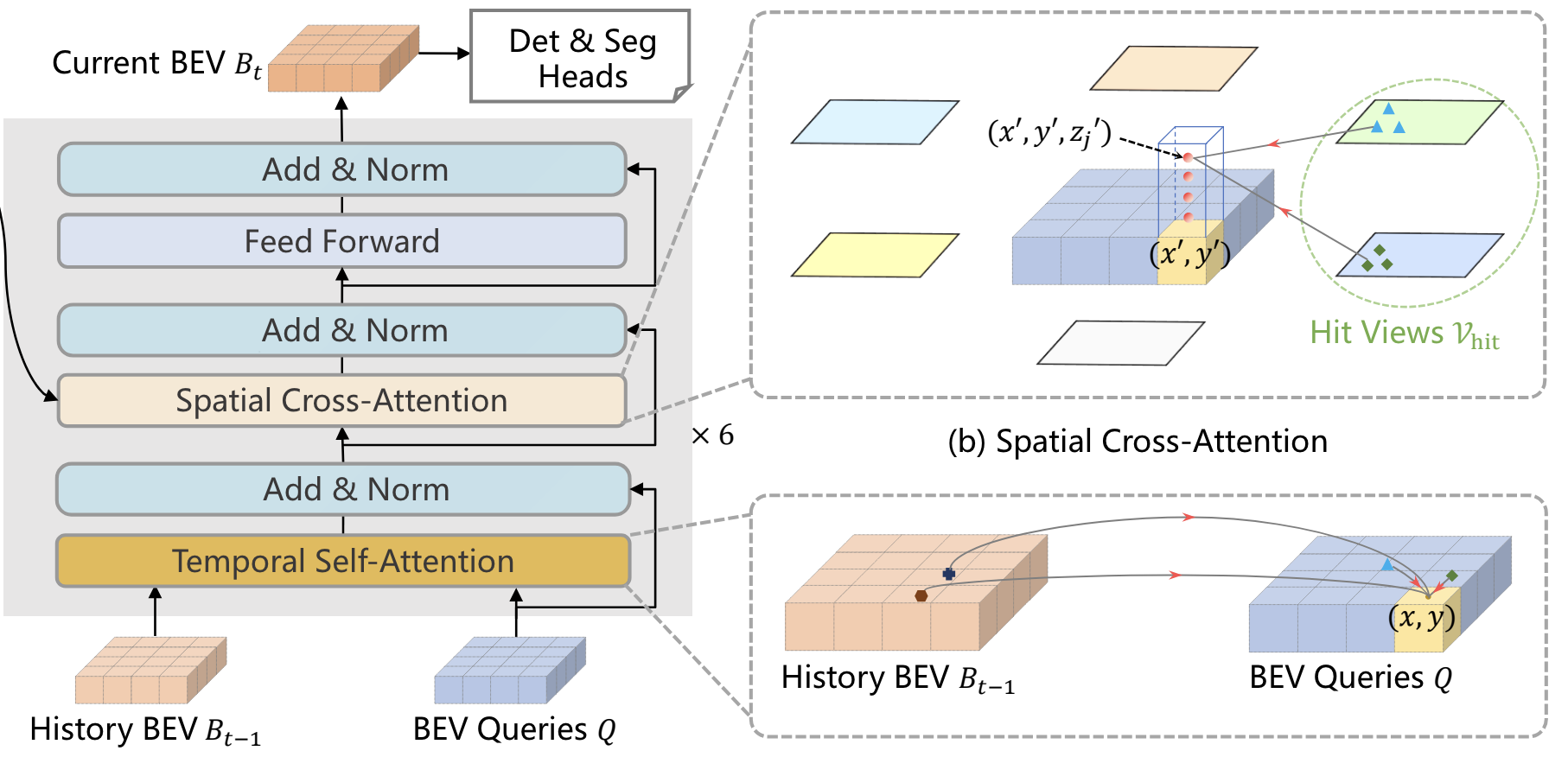}
    \caption{LaneSegNet Encoder Design (BEVFormer). \cite{lanesegnet} \cite{bevformer}}
    \label{fig:laneseg_encoder}
\end{figure}

The encoder's main purpose in transformer models is to process a typically large input sequence and compress the information into an encoded vector representation which contains the most relevant information from the input. The LaneSegNet encoder (Figure \ref{fig:laneseg_encoder}) is derived from the BEVFormer design \cite{bevformer}, whose novelty comes from its ability to transform PV features from a traditional feature extractor into a BEV representation. BEVFormer takes a proposal of BEV feature queries, and combines them with the previous time step's history BEV using the Temporal Self-Attention mechanism. This mechanism is derived from Deformable Attention, first introduced by Deformable-DETR \cite{deformable-detr}. Its purpose is to supply the network with context about the surrounding environment through combining its semantic understanding with temporal information. After Temporal Self-Attention, the BEVFormer encoder performs a Spatial Cross-Attention operation, also derived from Deformable Attention, which incorporates the temporally-enhanced BEV queries with the PV features from the feature extractor. This mechanism supplies the network with context about the surrounding environment through combining its perspective view with the temporal information computed in Temporal Self-Attention.

\begin{figure}[H]
    \centering
    \includegraphics[width=0.6\columnwidth]{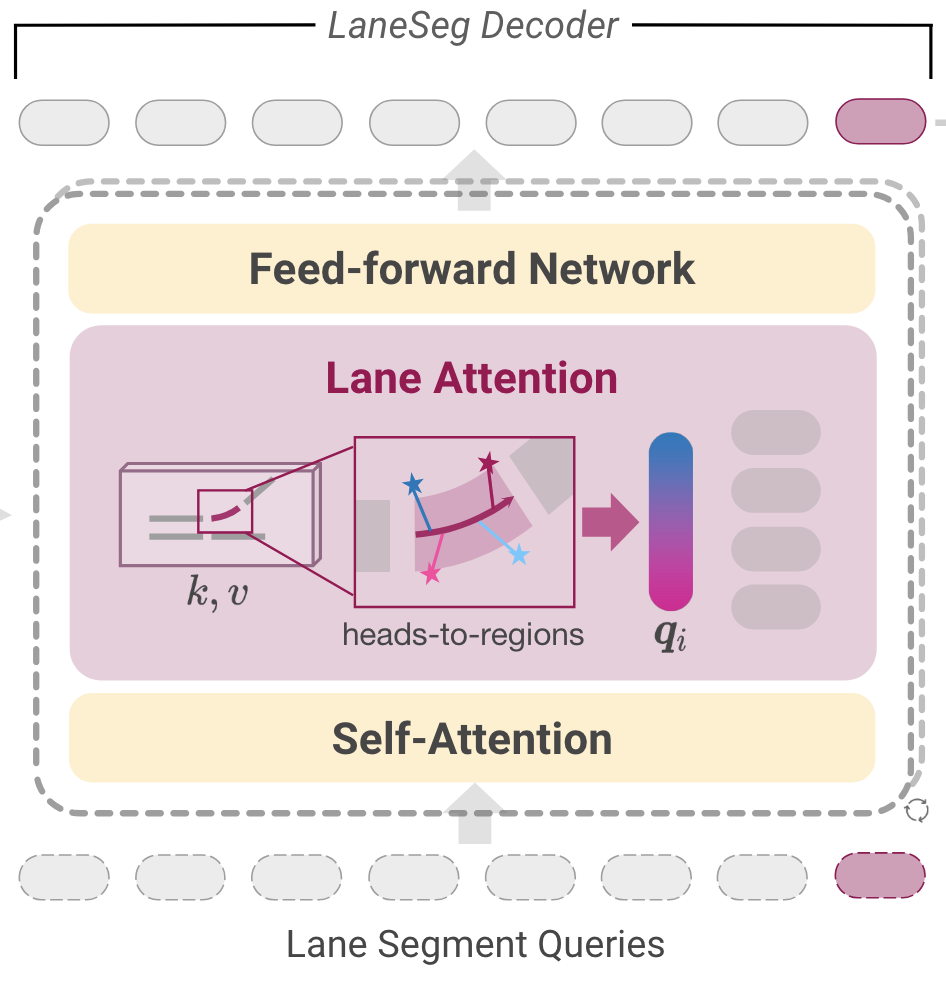}
    \caption{LaneSegNet Decoder Design. \cite{lanesegnet} \cite{deformable-detr}}
    \label{fig:laneseg_decoder}
\end{figure}

The decoder is responsible for taking the vector produced by the encoder and decoding it into a new semantic representation, depending on the desired output. In LaneSegNet, the decoder design (Figure \ref{fig:laneseg_decoder}) is based on the transformer used in Deformable-DETR \cite{deformable-detr}. This design takes in the encoded vector and then passes it through eight parallel self-attention heads and deformable attention mechanisms. The mechanisms are responsible for generating the lane line reference points and offsets that learn to converge around the correct shape for each line in the BEV representation of the road.

In most transformer designs, the encoder and decoder components are repeated over multiple layers. Increasing the number of layers in an encoder allows for increased representation capture, meaning that the model retrieves more abstract and complex patterns in the input data. For the decoder, having more layers allows it to deconstruct more complex output tasks such as object detection or classification. In practice, the decoder usually has an equal or greater depth than the encoder, depending on the output task. Increasing encoder and decoder layers allows the model to capture more complex representations in the input sequence, but increasing them too much will lead to model overfitting, as well as increased computation time.

In LaneSegNet, the authors propose three encoder layers and six decoder layers following the BEVFormer and Deformable-DETR designs, respectively. However, the authors do not give any specific reasons, or provide testing to demonstrate why they believed these hyperparameters were optimal. So, in an attempt to further understand the authors' intuition and potentially find settings that would allow for an acceptable tradeoff between time complexity and performance, the team investigated various combinations of layer counts for the LaneSegNet encoder and decoder stacks.

\newpage 
\section{\textbf{Results}}

\subsection{Model Replication}
Before any optimization methods were attempted, the team trained the LaneSegNet model for 30 epochs on our GPU, and achieved promising numerical and visual results. Although the official implementation was only trained for 24 epochs, our model was not quite able to surpass the accuracy of the official implementation. However, the accuracy metric of mean average-precision (mAP) and the visual outputs are are getting close to replicating the official results. After 30 epochs, our model produced a mAP of 23.5, which is quite a bit lower than the official implementation's mAP of 32.6. Table \ref{tab:lanesegresults} further contextualizes our metric results, with checkpoints at every multiple of 10 epochs.

\begin{table}[H]
    \large
    \centering
    \begin{tabular}{c|c|c}
    & Epochs & mAP \\
    \hline
    Official LaneSegNet & 24 & 32.6\\
    \hline
    Our LaneSegNet & 10 & 20.2\\
    Our LaneSegNet & 20 & 22.4\\
    Our LaneSegNet & 30 & 23.5\\
    \hline
    \end{tabular}
    \caption{Metric Comparison of LaneSegNet Implementation.}
    \label{tab:lanesegresults}
\end{table}

\begin{figure}[H]
    \centering
    \includegraphics[width=0.7\columnwidth]{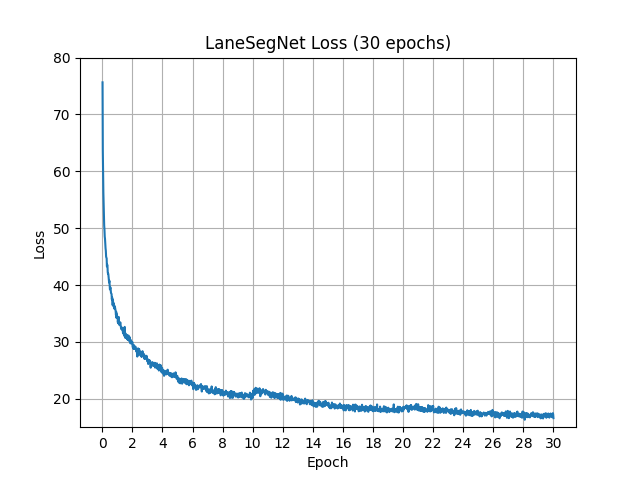}
    \caption{Our Network's Loss after 30 Epochs.}
    \label{fig:loss_curve_30e}
\end{figure}

The lower accuracy of our implementation can likely be attributed to computation constraints. Our model was not able to run more than 10 epochs at a time, so the training had to stopped and resumed at each multiple of 10 epochs. From the loss curve in Fig. \ref{fig:loss_curve_30e}, it appears that resuming training at these checkpoints causes noise to occur, where the loss jumps up before declining again. It is possible that resuming training like this is periodically stagnating the learning, and that a continuous training session may result in a smoother loss curve and higher mAP.

\begin{figure}[H]
    \centering
    \includegraphics[width=0.6\columnwidth]{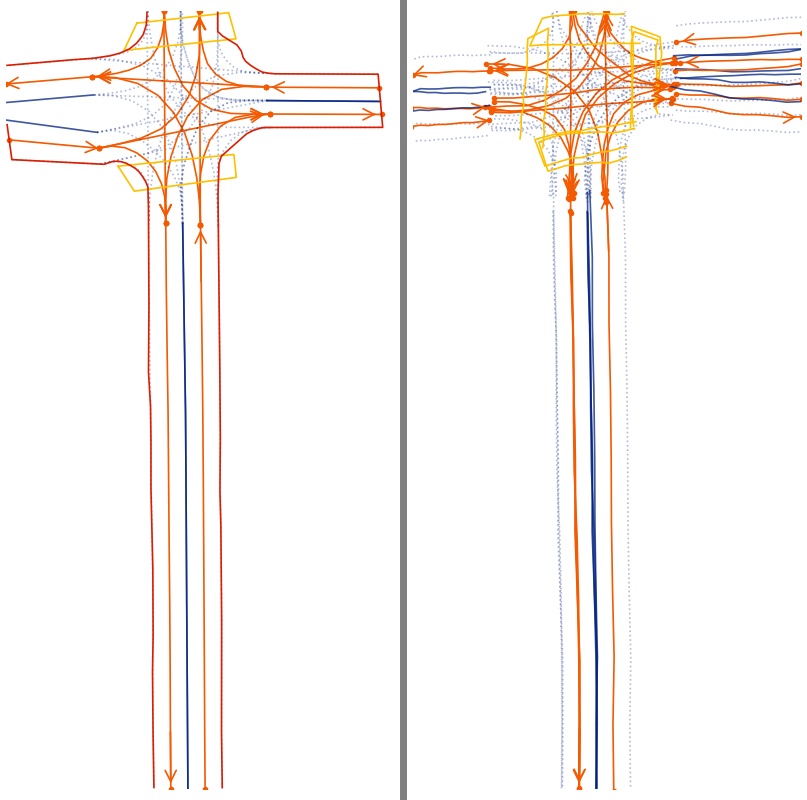}
    \caption{Groundtruth (Left) and Model Output (Right)}
    \label{fig:lanesegresults1}
\end{figure}

\begin{figure}[H]
    \centering
    \includegraphics[width=0.6\columnwidth]{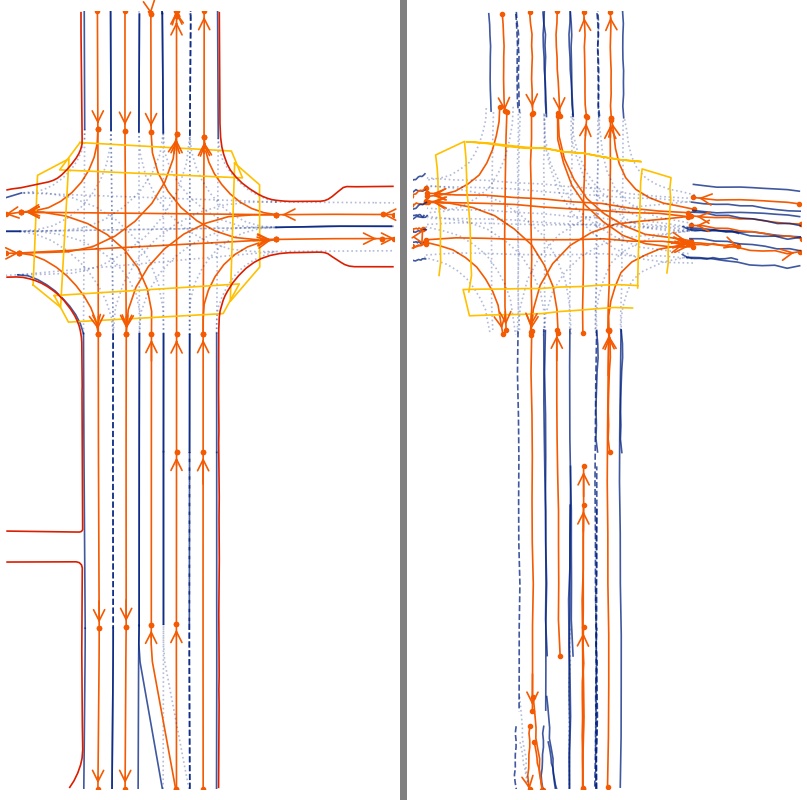}
    \caption{Groundtruth and Model Output}
    \label{fig:lanesegresults2}
\end{figure}

\begin{figure}[H]
    \centering
    \includegraphics[width=0.6\columnwidth]{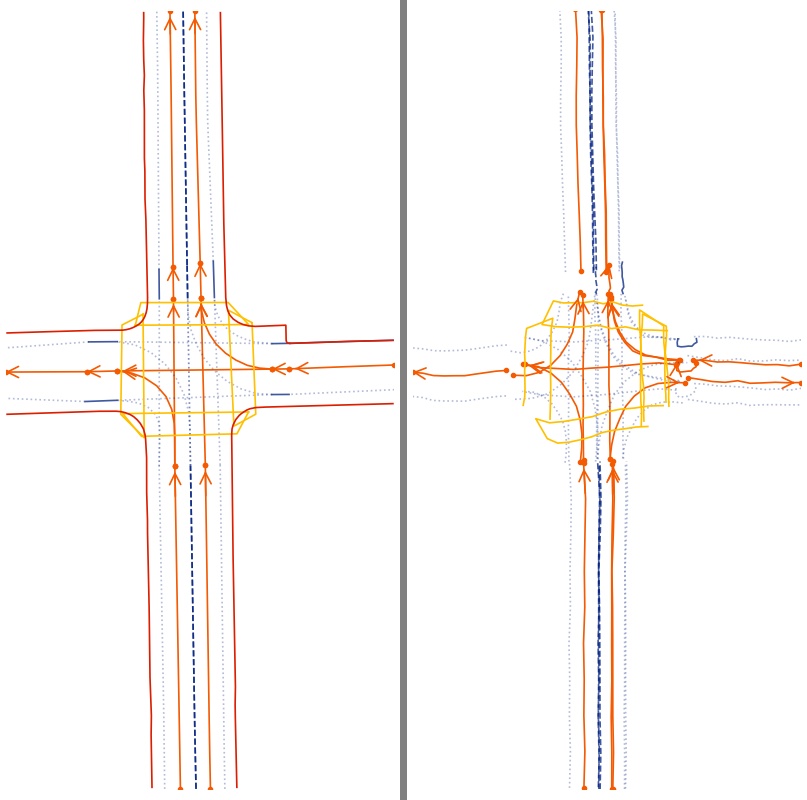}
    \caption{Groundtruth and Model Output}
    \label{fig:lanesegresults3}
\end{figure}

The results of the team's implementation can be further contextualized through a visualization of the outputs shown in Figures \ref{fig:lanesegresults1} - \ref{fig:lanesegresults3}. All these images are taken from the isolated test subset of the OpenLane-V2 dataset, for the purpose of measuring generalizability. As the visual results show, our model is able to capture the general structure of the road, and can accurately determine the locations of intersections. However, the model struggles to accurately predict the number and shapes of any crossing lanes in the perpendicular road.

\subsection{Optimization Results}

When training with our potential optimizations, the project team kept track of training time by measuring the average hours per epoch, and used this as a metric against the final mAP to determine how the changes affected the model's runtime and performance. Table \ref{tab:optimizations} illustrates the metric results for each optimization method. The first result showed that using a ResNet-18 feature extractor instead of ResNet-50 resulted in a slower training time and a $13.3\%$ decrease in mAP. Since no evaluation area showed promise, the team suspended training after 10 epochs. However the second and third results stemming from modifications in the encoder and decoder stacks offered interesting tradeoff alternatives in terms of training time and accuracy.

\begin{table}[H]
    \large
    \centering
    \begin{tabular}{c|c|c|c}
    & Epochs & hrs/Epoch & mAP \\
    \hline
    & 10 & 1.8 & 20.2\\
    Replicated LaneSegNet & 20 & 1.8 & 22.4\\
     & 30 & 1.8 & 23.5\\
    \hline
    ResNet-18 & 10 & 2 & 15.5\\
    2:4 E-D Ratio & 20 & 1.4 & 20.8\\
    4:8 E-D Ratio & 20 & 2 & 27.7\\
    \hline
    \end{tabular}
    \caption{Metric Comparison of LaneSegNet Optimizations.}
    \label{tab:optimizations}
\end{table}

Training with two encoder layers and four decoder layers reduced training time by $22.3\%$, with just a $7.1\%$ drop in accuracy after 20 epochs. As expected, making the encoder and decoder shallower proved to save time, at the cost of some accuracy. Furthermore, training with four encoder layers and eight decoder layers increased training time by only $11.1\%$, with an astonishing $23.7\%$ increase in accuracy after 20 epochs, outperforming even our unmodified model after 30 epochs.

\begin{figure}[H]
    \centering
    \includegraphics[width=0.6\columnwidth]{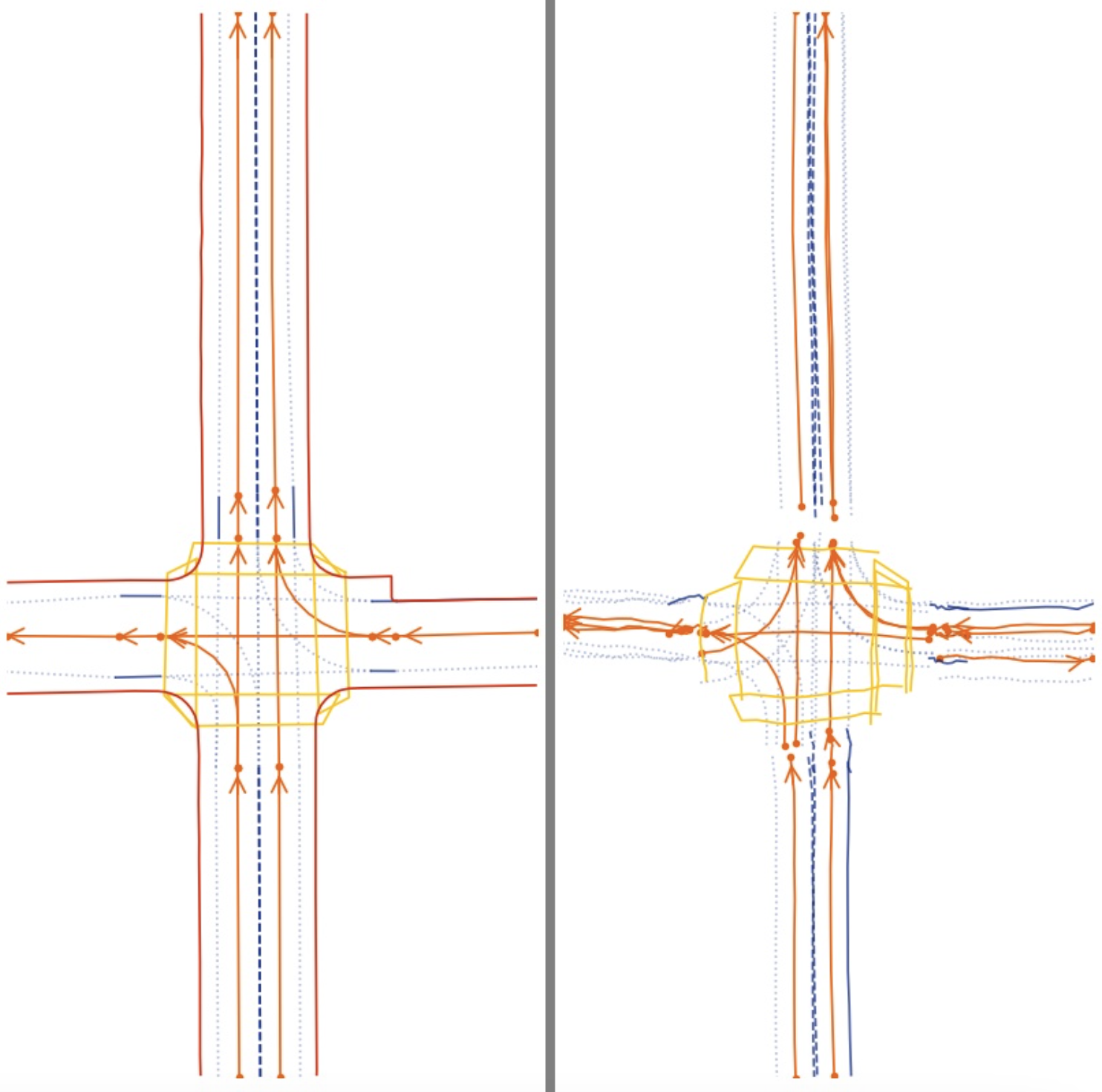}
    \caption{Output without Optimizations}
    \label{fig:optimresults1}
\end{figure}

\begin{figure}[H]
    \centering
    \includegraphics[width=0.6\columnwidth]{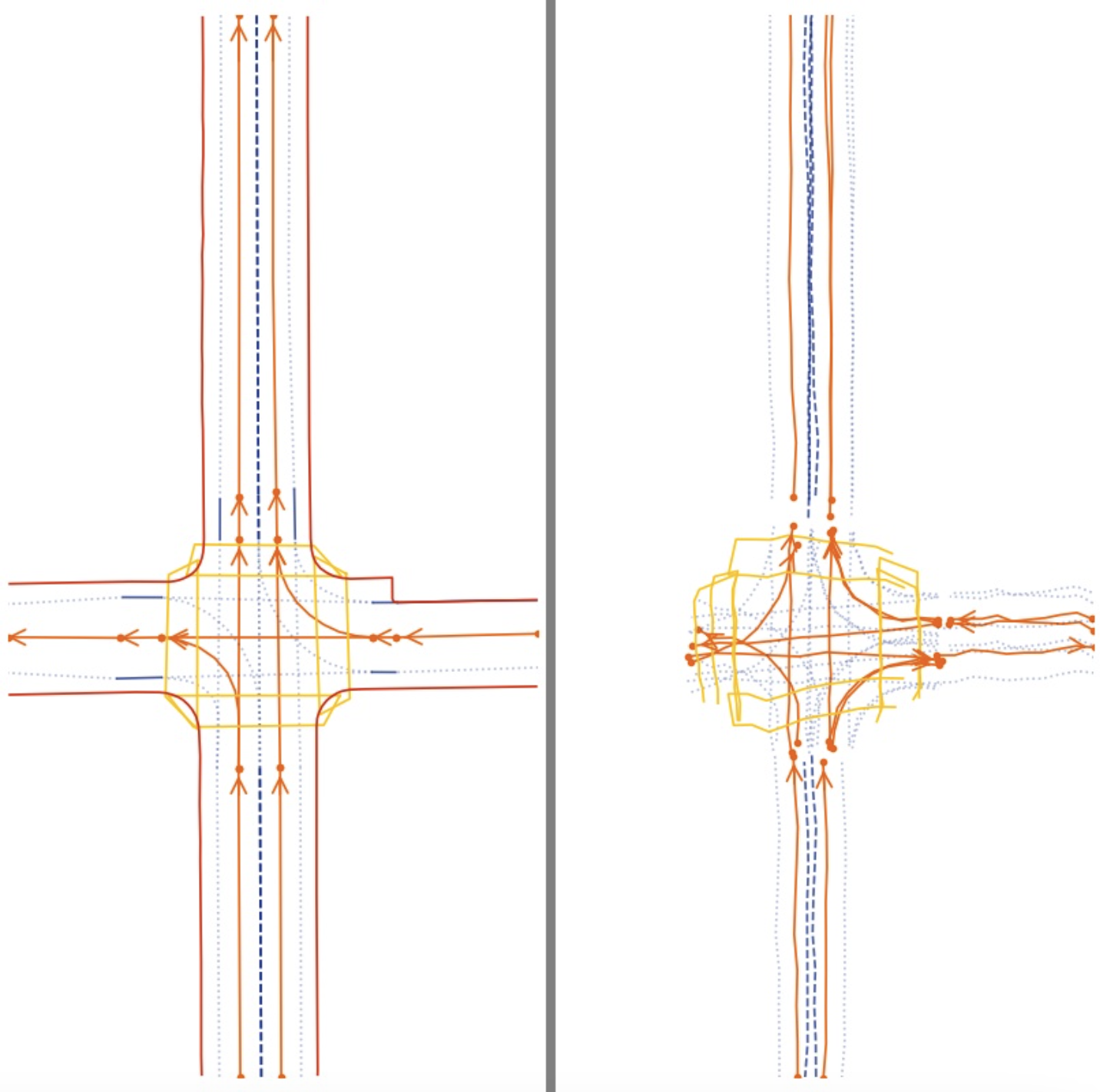}
    \caption{Output with ResNet-18}
    \label{fig:optimresults2}
\end{figure}

\begin{figure}[H]
    \centering
    \includegraphics[width=0.6\columnwidth]{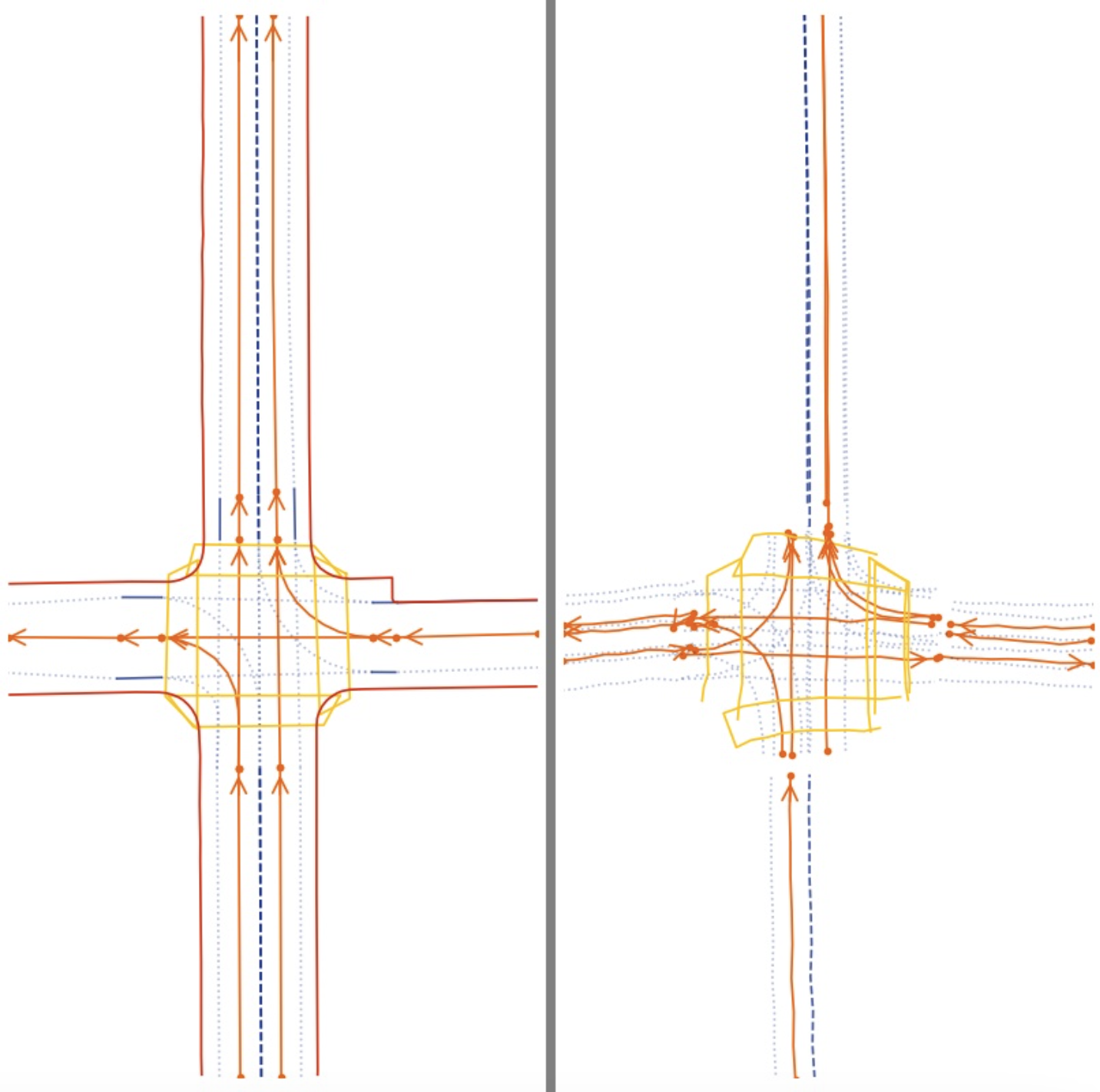}
    \caption{Output with 2:4 Encoder-Decoder Stacks}
    \label{fig:optimresults3}
\end{figure}

\begin{figure}[H]
    \centering
    \includegraphics[width=0.6\columnwidth]{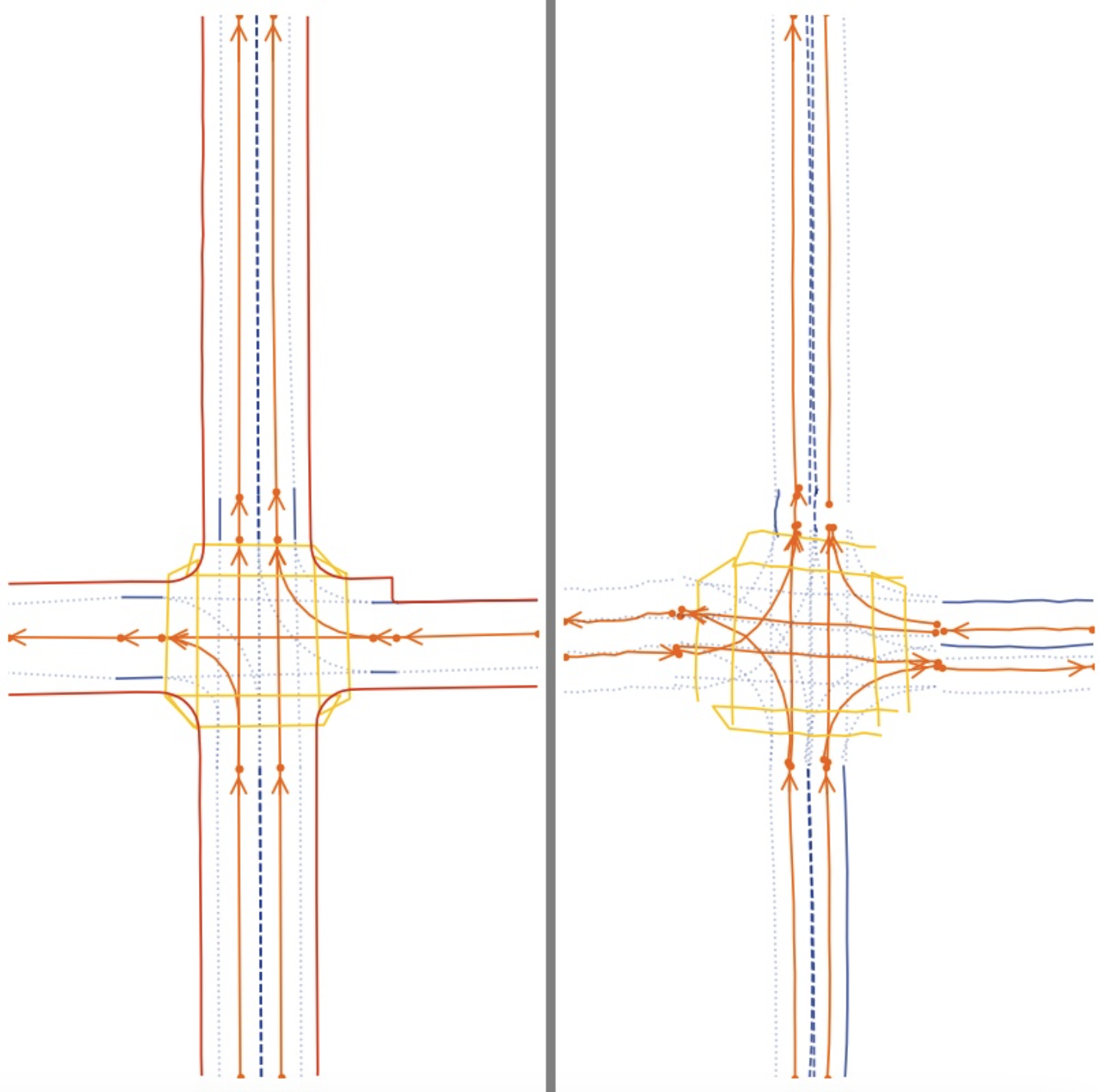}
    \caption{Output with 4:8 Encoder-Decoder Stacks}
    \label{fig:optimresults4}
\end{figure}

The results of these optimizations can be further contextualized in Figures \ref{fig:optimresults1} - \ref{fig:optimresults4}. All these images were taken from the same test example. As expected, the ResNet-18 optimization and 2:4 stack structure have some glaring differences with the groundtruth, and the unmodified implementation and 4:8 stack provide a much smoother and more accurate prediction.

\subsection{Conclusion}

The team found that the ResNet-18 optimization was not beneficial to the model in any way, as it increased the training time without any increase in accuracy. Even though ResNet-18 was shown to be much faster than ResNet-50, with just a slight drop in accuracy on ImageNet, it is likely that the feature extractor is much less significant in the entire LaneSegNet architecture than originally thought, thus would explain why improvements were seen in training time. It is also possible that the incomplete features produced by a shallower feature extractor provided the encoder with a less sensible semantic understanding, thus placing more strain on the decoder when calculating lane reference points. This would explain why training time actually increased with ResNet-18, when in theory it should have decreased.

As for the encoder and decoder stack modifications, the team found much more promising results. As we predicted, shallower stacks meant less computation in the more prominent components of the overall architecture, resulting in a $23.7\%$ reduction in training time. The expected drop in accuracy was only $7.1\%$, suggesting that the model did not suffer from underfitting with the shallower stacks, and could still provide reliable output. With a deeper stack structure of four encoders and eight decoders, it was surprising to see that the training time did not increase by much - only $11.1\%$. Even more surprising was the large increase in accuracy - arriving at a mAP of 27.7 after 20 epochs was unexpected. This was especially surprising considering how our replicated implementation stopped at 23.5 after 30 epochs. To put this result in context with the official implemention: our optimization obtained a mAP of 27.7 after 20 epochs, whereas the official implementation achieved 32.6 after 24 epochs with six times as much computation power in their GPUs. It would be interesting to see what accuracy the original authors could achieve if they used a 4:8 stack structure with their eight powerful GPUs.

In closing, we believe to have found two optimizations to the LaneSegNet architecture, through hyperparameter tuning, that could help others in their use of the model. For users with computation constraints, a 2:4 stack structure can be employed for a promising reduction in training time without much loss in accuracy. For users with access to high-powered GPUs, it is possible that using a 4:8 stack structure would provide higher accuracy, even surpassing that of the original implementation, with a negligible increase in training time.

\pagebreak
\bibliography{reference}
\bibliographystyle{IEEEtran}
\end{document}